\pdfoutput=1

\documentclass[11pt]{article}

\usepackage[]{EMNLP2022}

\usepackage{times}
\usepackage{latexsym}
\usepackage{algorithm}
\usepackage{algorithmic}
\usepackage[switch]{lineno} 
\usepackage[utf8]{inputenc}
\usepackage{CJKutf8}
\usepackage{multirow}
\usepackage{booktabs}
\usepackage{graphicx}
\usepackage{amsmath}
\usepackage{amssymb}
\usepackage{xpinyin}
\usepackage{pifont}
\usepackage{subfigure}

\newcommand{\MethodName}{\textsc{LeaD}}

\definecolor{jiColor}{RGB}{31,119,180}
\definecolor{zhiColor}{RGB}{255,127,14}
\definecolor{xinColor}{RGB}{0,128,0}
\definecolor{yingColor}{RGB}{255,0,0}

\usepackage[T1]{fontenc}

\usepackage{microtype}

\usepackage{inconsolata}

\title{Learning from the Dictionary: Heterogeneous Knowledge \\ Guided Fine-tuning for Chinese Spell Checking}

\author{Yinghui Li$^{1}$\thanks{ $^*$ indicates equal contribution. Work is done during Yinghui's internship at Tencent Cloud Xiaowei.},~Shirong Ma$^{1*}$,~Qingyu Zhou$^{2*}$,~Zhongli Li$^{2}$,~Yangning Li$^{1}$,\\ ~\textbf{Shuling Huang}$^{1}$,~\textbf{Ruiyang Liu}$^{5}$,~\textbf{Chao Li}$^{4}$,~\textbf{Yunbo Cao}$^{2}$ \and \textbf{Hai-Tao Zheng}$^{1,3}$\thanks{ $^{\dagger}$ Corresponding author: Hai-Tao Zheng. (E-mail: zheng.haitao@sz.tsinghua.edu.cn)} \\
        $^{1}$Tsinghua Shenzhen International Graduate School, Tsinghua University \\ 
        $^{2}$Tencent Cloud Xiaowei, $^{3}$Peng Cheng Laboratory , $^{4}$Xiaomi Group \\
        $^{5}$Department of Computer Science and Technology, Tsinghua University\\
        \texttt{\{liyinghu20, masr21\}@mails.tsinghua.edu.cn},
        \texttt{qingyuzhou@tencent.com}}

\begin{document}
\maketitle
\begin{abstract}
Chinese Spell Checking (CSC) aims to detect and correct Chinese spelling errors.
Recent researches start from the pretrained knowledge of language models and take multimodal information into CSC models to improve the performance. 
However, they overlook the rich knowledge in the dictionary, the reference book where one can learn how one character should be pronounced, written, and used.
In this paper, we propose the \MethodName{} framework, which renders the CSC model to \textbf{lea}rn heterogeneous knowledge from the \textbf{d}ictionary in terms of phonetics, vision, and meaning.
\MethodName{} first constructs positive and negative samples according to the knowledge of character phonetics, glyphs, and definitions in the dictionary. 
Then a unified contrastive learning-based training scheme is employed to refine the representations of the CSC models.
Extensive experiments\footnote{The source codes are available at \url{https://github.com/geekjuruo/LEAD}.} and detailed analyses on the SIGHAN benchmark datasets demonstrate the effectiveness of our proposed methods.
\end{abstract}

\section{Introduction}
\label{sec:Intro}

\begin{CJK*}{UTF8}{gbsn}
\begin{table}[t]
\centering
\small
\renewcommand\arraystretch{1.2}
\begin{tabular}{|r p{0.65\columnwidth}|}
\hline
\multicolumn{2}{|c|}{\textbf{\emph{Phonetically Similar Error}}} \\
\hline
\emph{Input} &  \textcolor{orange}{铁轨}上有一列\textcolor{red}{或(\textit{\pinyin{huo4})}}车在行驶。\\
\emph{Candidate 1} & \textcolor{orange}{铁轨}上有一列\textcolor{purple}{货(\textit{\pinyin{huo4})}}车在行驶。\\
& There is a \textcolor{purple}{\textit{truck}} running on the \textcolor{orange}{railway}.\\
\emph{Candidate 2} & \textcolor{orange}{铁轨}上有一列\textcolor{blue}{火(\textit{\pinyin{huo3})}}车在行驶。\\
& There is a \textcolor{blue}{\textit{train}} running on the \textcolor{orange}{railway}.\\
\emph{Definition} & 【\textcolor{blue}{火}车】一种交通工具，由机车牵引若干节车厢在\textcolor{orange}{铁路}上行驶。 \\
 & A means of transportation in which a number of carriages are pulled by a locomotive to travel on a \textcolor{orange}{railway}. \\
\hline
\multicolumn{2}{|c|}{\textbf{\emph{Visually Similar Error}}} \\
\hline
\emph{Input} &  \textcolor{orange}{炉子}上正\textcolor{red}{绕(\textit{\pinyin{rao4})}}着水。\\
\emph{Candidate 1} & \textcolor{orange}{炉子}上正\textcolor{purple}{浇(\textit{\pinyin{jiao1})}}着水。\\
& Water is \textcolor{purple}{\textit{pouring}} on the \textcolor{orange}{stove}.\\
\emph{Candidate 2} & \textcolor{orange}{炉子}上正\textcolor{blue}{烧(\textit{\pinyin{shao1})}}着水。\\
& Water is \textcolor{blue}{\textit{burning}} on the \textcolor{orange}{stove}.\\
\emph{Definition} & 【\textcolor{blue}{烧}】\textcolor{orange}{加热}使物体发生变化。 \\
 & Change matters by \textcolor{orange}{heating}. \\
\hline
\end{tabular}

\caption{Examples of Chinese spelling errors. The \textcolor{red}{wrong}/\textcolor{purple}{candidate}/\textcolor{blue}{golden} characters are in \textcolor{red}{red}/\textcolor{purple}{purple}/\textcolor{blue}{blue}. \textcolor{orange}{The key information} is in \textcolor{orange}{orange}.}
\label{tab:Intro_Table}
\end{table}

As a crucial Chinese processing task, Chinese Spell Checking (CSC) aims to detect and correct Chinese spelling errors~\cite{wu-etal-2013-integrating}, which are mainly caused by phonetically or visually similar characters~\cite{liu-etal-2010-visually}.
Recent researches propose to introduce phonetics and vision information to help pretrained language models (PLMs) deal with confusing characters~\cite{liu-etal-2021-plome, xu-etal-2021-read, huang-etal-2021-phmospell}. 
However, CSC is challenging because it requires not only phonetics/vision information but also complex definition knowledge to assist in finding the truly correct character. 
As shown in Table~\ref{tab:Intro_Table}, the “货(\textit{\pinyin{huo4}})” and “火(\textit{\pinyin{huo3}})” are phonetically similar, and both are suitable collocations with “车”. But if the model pays attention to the keyword “铁轨(railway)” and knows the meaning of the “火车(train)”, then the model can not be disturbed by the “货” and easily make the correct judgment. The same situation also occurs in the visual case.
For these hard samples, PLMs do not perform well in that the masked-language modeling objective determines their pretrained semantic knowledge is more about the collocation of characters, rather than the definitions of their meanings.
Therefore, if the model understands the word meanings, it can be further enhanced to handle more hard samples and get performance improvements.

\end{CJK*}

To help people learn Chinese, the meanings of Chinese characters and words have been pre-organized as the definition sentences in the dictionary.
The dictionary contains a wealth of useful knowledge for CSC, including character phonetics, glyphs, and definitions. 
It is also the most important resource for Chinese beginners to learn how to pronounce, write, and use one character. 
Inspired by this, we focus on utilizing the rich knowledge in the dictionary to improve the CSC performance.

In this paper, we propose \MethodName{}, a unified fine-tuning framework to guide the CSC models to \textbf{lea}rn heterogeneous knowledge from the \textbf{d}ictionary. 
In general, \MethodName{} has one training paradigm but three different training objectives besides the traditional CSC objective.
This enables models to learn three different kinds of knowledge, namely phonetics, vision, and definition knowledge.
Specifically, we construct various positive and negative samples according to the respective characteristics of different knowledge, and then utilize these generated sample pairs to train models with our designed unified contrastive learning paradigm. 

Through the optimization of \MethodName{}, the fine-tuned model handles various phonetically/visually similar character errors as well as previous multimodal models, and goes a further step to deal with more confusing errors with the help of the definition knowledge contained in the dictionary.
Additionally, \MethodName{} is a model-agnostic fine-tuning framework, which has no restrictions on the fine-tuned models. 
In practice, we fine-tune BERT and a more complex multimodal CSC model~\citep{xu-etal-2021-read} with \MethodName{}, and experimental results on the SIGHAN datasets show consistent improvements.

To summarize, the contributions of our work are in three folds: 
(1) We focus on the importance of the dictionary knowledge for the CSC task, which is instructive for future CSC research. 
(2) We propose the \MethodName{} framework, which fine-tunes the models to learn heterogeneous knowledge beneficial to the CSC task in a unified manner. 
(3) We conduct extensive experiments and detailed analyses on widely used SIGHAN datasets and \MethodName{} outperforms previous state-of-the-art methods.

\section{Related Work}
\subsection{Chinese Spell Checking}
Recently, deep learning-based models have gradually become the mainstream CSC methods~\cite{wang-etal-2018-hybrid, hong-etal-2019-faspell, zhang-etal-2020-spelling, li-etal-2022-past}.
SpellGCN~\cite{cheng-etal-2020-spellgcn} uses GCN~\cite{kipf2017semi} to fuse character embedding with similar pronunciation and shape, explicitly modeling the relationship between characters.
GAD~\cite{guo-etal-2021-global} proposes a global attention decoder method and pre-trains the BERT~\cite{devlin-etal-2019-bert} with a confusion set guided replacement strategy.
\citet{li-etal-2021-exploration} proposes a method that continually identifies the weak spots of a model to generate more valuable training samples, and applies a task-specific pre-training strategy to enhance the model.
Additionally, many CSC works have focused on the importance of multimodal knowledge for CSC.
DCN~\cite{wang-etal-2021-dynamic}, MLM-phonetics~\cite{zhang-etal-2021-correcting}, and SpellBERT~\cite{ji-etal-2021-spellbert} all utilize phonetic features to improve CSC performance.
PLOME~\cite{liu-etal-2021-plome} designs a confusion set-based masking strategy and introduces phonetics and stroke information.
REALISE~\cite{xu-etal-2021-read} and PHMOSpell~\cite{huang-etal-2021-phmospell} both employ kinds of encoders to learn multimodal knowledge. 
Different from previous works, our work is the first to introduce definition knowledge from the dictionary to enhance CSC models.

\subsection{Contrastive Learning}
Contrastive learning is a kind of representation learning method that has been widely used in NLP and CV~\cite{chen2020simple, he2020momentum, gao2021simcse}.
The main motivation of contrastive learning is to attract the positive samples and repulse the negative samples in a certain space~\cite{HadsellCon, chen2020simple, khosla2020supervised}. 
In the NLP field, various contrastive learning methods have been studied for learning all kinds of better representations, such as entity~\cite{li2022contrastive}, sentence~\cite{kim-etal-2021-self}, and relation~\cite{qin-etal-2021-erica}.
To the best of our knowledge, we are the first to leverage the idea of contrastive learning to learn better phonetics, vision, and definition knowledge for CSC.

\begin{figure*}[]
\centering
\includegraphics[height=0.51\textwidth]{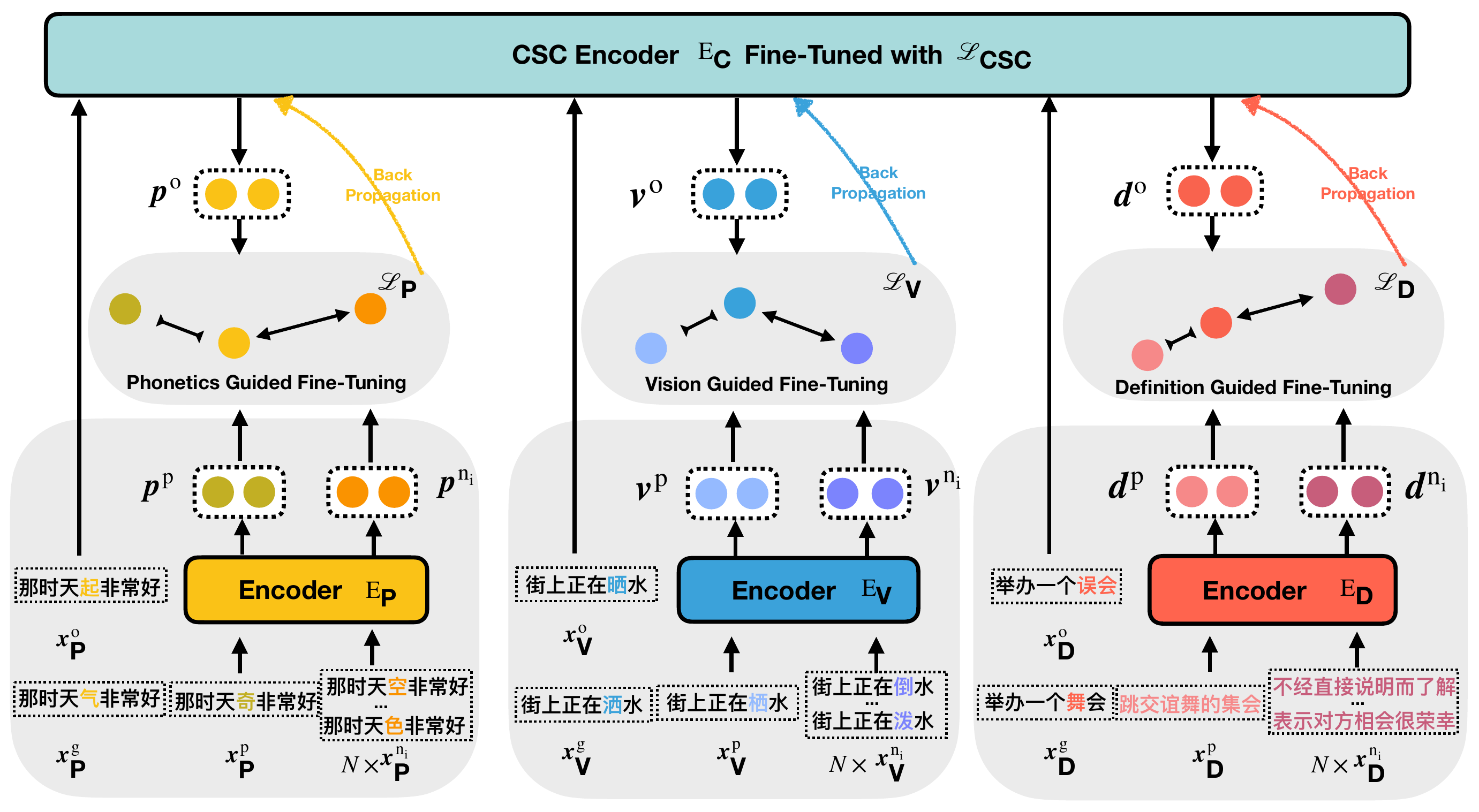}
\caption{Overview of our \MethodName{} framework. According to the contrastive learning mechanism proposed in~\cite{DBLP:conf/cvpr/He0WXG20}, the gradients of $\mathcal{L}_{\text{P}}, \mathcal{L}_{\text{V}}, \mathcal{L}_{\text{D}}$ are propagated back to the CSC model so that it is optimized accordingly. }
\label{Pos_Neg_Figure}
\end{figure*}

\section{Methodology}
In this section, we first introduce the overview of the \MethodName{} framework, as illustrated in Figure~\ref{Pos_Neg_Figure}, and describe our designed unified contrastive learning mechanism for heterogeneous dictionary knowledge.
Then, for each knowledge-guided fine-tuning, we explain its motivation, positive/negative pairs construction, and representation metric which is used in the contrastive learning mechanism.

\subsection{Overview of \MethodName{}}
\label{Method_Overview}
In \MethodName{}, in addition to using the CSC samples to train the traditional CSC objective, various positive and negative pairs are generated for the contrastive learning of three kinds of knowledge (i.e., phonetics, vision, and definition). 
Specifically, for a particular knowledge $\text{K}$, to achieve a training mini-batch, we construct a positive pair $(\boldsymbol{x}_{\text{K}}^{\mathrm{o}}, \boldsymbol{x}_{\text{K}}^{\mathrm{p}})$ and $N$ negative pairs $\{ (\boldsymbol{x}_{\text{K}}^{\mathrm{o}}, \boldsymbol{x}_{\text{K}}^{\mathrm{n_i}}) \}_{i=0}^{N-1}$, where $\text{K} \in \{\text{P}, \text{V}, \text{D} \}$ represents “\textbf{P}honetics, \textbf{V}ision, \textbf{D}efinition” knowledge.
Note that the original sample $\boldsymbol{x}_{\text{K}}^{\mathrm{o}}$ is directly from the CSC samples, the positive sample $\boldsymbol{x}_{\text{K}}^{\mathrm{p}}$ and negative samples $\{\boldsymbol{x}_{\text{K}}^{\mathrm{n_i}}\}$ are generated from $\boldsymbol{x}_{\text{K}}^{\mathrm{o}}$ according to the characteristics of the knowledge $\text{K}$.

Then, for the positive and negative sentences (i.e., $\boldsymbol{x}_{\text{K}}^{\mathrm{p}}$ and $\{\boldsymbol{x}_{\text{K}}^{\mathrm{n_i}}\}$) of length $T$, we use various encoders (i.e., $\mathrm{E}_{\text{K}} \in \{\mathrm{E}_{\text{P}}, \mathrm{E}_{\text{V}}, \mathrm{E}_{\text{D}} \}$) to map them to a sequence of representations  $\boldsymbol{k}^{\mathrm{p}} = [k_1^{\mathrm{p}}, ..., k_T^{\mathrm{p}}], \{\boldsymbol{k}^{\mathrm{n_i}}\} = \{[k_1^{\mathrm{n_i}}, ..., k_T^{\mathrm{n_i}}]\}, k_j^{\mathrm{p}}, k_j^{\mathrm{n_i}}\in \mathbb{R}^{h}$, where $h$ is the size of the $\mathrm{E}_{\text{K}}$'s hidden state:
\begin{gather}
  \boldsymbol{k}^{\mathrm{p}} = \mathrm{E}_{\text{K}}(\boldsymbol{x}_{\text{K}}^{\mathrm{p}}), \boldsymbol{k}^{\mathrm{p}} \in \{\boldsymbol{p}^{\mathrm{p}}, \boldsymbol{v}^{\mathrm{p}}, \boldsymbol{d}^{\mathrm{p}} \}, \\
  \{\boldsymbol{k}^{\mathrm{n_i}}\} = \{\mathrm{E}_{\text{K}}(\boldsymbol{x}_{\text{K}}^{\mathrm{n_i}})\}, 
    \boldsymbol{k}^{\mathrm{n_i}} \in \{ \boldsymbol{p}^{\mathrm{n_i}}, \boldsymbol{v}^{\mathrm{n_i}}, \boldsymbol{d}^{\mathrm{n_i}}\}.
\end{gather}
For the original sentence $\boldsymbol{x}_{\text{K}}^{\mathrm{o}}$, we utilize the encoder of CSC model (i.e., $\mathrm{E}_{\text{C}}$) to get its sentence representation $\boldsymbol{k}^{\mathrm{o}} = [k_1^{\mathrm{o}}, ..., k_T^{\mathrm{o}}], k_j^{\mathrm{o}} \in \mathbb{R}^{h}$, the $\mathrm{E}_{\text{C}}$'s hidden size is equal the dimension of the $\mathrm{E}_{\text{K}}$'s hidden state:
\begin{equation}
    \boldsymbol{k}^{\mathrm{o}} = \mathrm{E}_{\text{C}}(\boldsymbol{x}_{\text{K}}^{\mathrm{o}}),
    \boldsymbol{k}^{\mathrm{o}} \in \{\boldsymbol{p}^{\mathrm{o}}, \boldsymbol{v}^{\mathrm{o}}, \boldsymbol{d}^{\mathrm{o}} \}.
\end{equation}

After obtaining the representations of our generated sentence pairs, following the widely used InfoNCE~\cite{DBLP:journals/corr/abs-1807-03748}, we train these sample pairs in a contrastive manner:
\begin{equation}
\mathcal{L}_{\text{K}} = - \log \frac{f_\text{K}(\boldsymbol{k}^{\mathrm{o}}, \boldsymbol{k}^{\mathrm{p}}, s)}{f_\text{K}(\boldsymbol{k}^{\mathrm{o}}, \boldsymbol{k}^{\mathrm{p}}, s)+ \sum\limits_{i=0}^{N-1} f_\text{K}(\boldsymbol{k}^{\mathrm{o}}, \boldsymbol{k}^{\mathrm{n_i}}, s)},
\end{equation}
where the $\mathcal{L}_{\text{K}}$ is the training objective of the knowledge $\text{K}$, and the $f_\text{K}$ is the representation metric function in the respective space of each knowledge, which will be introduced in later sections. In the mini-batch, all sentences are of length $T$ and their $s$-th character is the spelling error.

It is worth emphasizing that the three knowledge encoders (i.e., $\mathrm{E}_{\text{P}}$, $\mathrm{E}_{\text{V}}$, and $\mathrm{E}_{\text{D}}$) are frozen, while the $\mathrm{E}_{\text{C}}$ receives gradients from multiple dimensions and is optimized during the training process. 
Besides, our proposed \MethodName{} is model-agnostic so that we can arbitrarily configure $\mathrm{E}_{\text{P}}$, $\mathrm{E}_{\text{V}}$, $\mathrm{E}_{\text{D}}$ and easily use previous CSC models as $\mathrm{E}_{\text{C}}$. The implementation details of various encoders in our experiments are shown in Appendix~\ref{sec:implementation_appendix}.

Briefly, our proposed \MethodName{} performs specific contrastive fine-tuning guided by heterogeneous knowledge, thereby introducing various beneficial information into CSC models to improve their performance. 
In the Sections~\ref{Method_PCL}-~\ref{Method_DCL}, we will detail the positive and negative pairs construction and representation metric we design for each knowledge.

\subsection{Phonetics Guided Fine-tuning}
\label{Method_PCL}
According to the phonetics knowledge, Chinese characters are represented by Pinyin. Therefore, to make the model better handle phonetic errors, we aim to guide it to pay more attention to characters with similar Pinyin. To this end, we propose the \emph{Phonetics Guided Fine-tuning}, which aims to refine the representation space learned by models so that the representations of the similar Pinyin characters are pulled closer while the representations of different Pinyin characters are pushed outward. Thus, when handling phonetically spelling errors, our model will preferentially associate with their corresponding phonetically similar characters.

\paragraph{Positive and Negative Pairs}
\begin{CJK*}{UTF8}{gbsn}
For the phonetics knowledge, we regard characters with similar Pinyin as positive pairs and characters with different Pinyin as negative pairs.
As shown in Figure~\ref{Pos_Neg_Figure}, given a training sample $\boldsymbol{x}_{\text{P}}^{\mathrm{o}}$ “那时天\underline{起(\pinyin{qi3}, rise)}非常好” that has a phonological spelling error, we replace “起(\pinyin{qi3}, rise)” with its phonetically similar character “奇(\pinyin{qi2}, strange)” to achieve a positive sample $\boldsymbol{x}_{\text{P}}^{\mathrm{p}}$. To generate negative samples $\{\boldsymbol{x}_{\text{P}}^{\mathrm{n_i}}\}$, we randomly select $N$ characters with different Pinyin, such as “色(\pinyin{se4}, color)”, to replace “起(\pinyin{qi3}, rise)”. Finally, we will get a positive pair $(\boldsymbol{x}_{\text{P}}^{\mathrm{o}}, \boldsymbol{x}_{\text{P}}^{\mathrm{p}})$ and $N$ negative pairs $\{(\boldsymbol{x}_{\text{P}}^{\mathrm{o}}, \boldsymbol{x}_{\text{P}}^{\mathrm{n_i}})\}$ to form a mini-batch for the fine-tuning of phonetics knowledge.
\end{CJK*}

\paragraph{Representation Metric}
Note that the motivation of phonetics guided fine-tuning is to refine the character-level representation of CSC models under the constraints of phonetics knowledge, so we only need the representation of the spelling error position, i.e., the $s$-th character.
Therefore, the representation metric of phonetics guided fine-tuning (i.e., $f_{\text{P}}$) is calculated as the dot product function:
\begin{gather}
    f_{\text{P}}(\boldsymbol{p}^{\mathrm{o}}, \boldsymbol{p}^{\mathrm{p}}, s) = \text{exp}(p_{s}^{\mathrm{o}\top}p_{s}^{\mathrm{p}}), \\
    f_{\text{P}}(\boldsymbol{p}^{\mathrm{o}}, \boldsymbol{p}^{\mathrm{n_i}}, s) = \text{exp}(p_{s}^{\mathrm{o}\top}p_{s}^{\mathrm{n_i}}).
\end{gather}

\subsection{Vision Guided Fine-tuning}
\label{Method_VCL}
Similar to the phonetics guided fine-tuning, we propose the \emph{Vision Guided Fine-tuning} for better vision representations and improving the visual error correction ability of models. Specifically, based on the fact that Chinese characters are composed of strokes in the dimension of vision knowledge, the purpose of this module is to train models to represent characters with more similar strokes closer and characters with more different strokes farther away in the visual representation space.

\paragraph{Positive and Negative Pairs}
\begin{CJK*}{UTF8}{gbsn}
Based on the visual similarity between characters, for a specific Chinese character, we directly obtain its characters with similar strokes from the pre-defined confusion set widely used in previous works~\cite{wang-etal-2019-confusionset,cheng-etal-2020-spellgcn,zhang-etal-2020-spelling}. 
Take Figure~\ref{Pos_Neg_Figure} as an example, for a training sample $\boldsymbol{x}_{\text{V}}^{\mathrm{o}}$ “街上正在\underline{晒(\pinyin{shai4}, bask)}水”, its positive sample $\boldsymbol{x}_{\text{V}}^{\mathrm{p}}$ is generated by replacing “晒(\pinyin{shai4}, bask)” with “栖(\pinyin{qi1}, habitat)”. Similar to the phonetics guided fine-tuning, characters with different strokes are randomly selected to generate the $\{\boldsymbol{x}_{\text{V}}^{\mathrm{n_i}}\}$.
\end{CJK*}

\paragraph{Representation Metric}
Similar to the $f_{\text{P}}$, we also utilize the dot product metric to measure the representation distance in vision space:
\begin{gather}
    f_{\text{V}}(\boldsymbol{v}^{\mathrm{o}}, \boldsymbol{v}^{\mathrm{p}}, s) = \text{exp}(v_{s}^{\mathrm{o}\top}v_{s}^{\mathrm{p}}),\\
    f_{\text{V}}(\boldsymbol{v}^{\mathrm{o}}, \boldsymbol{v}^{\mathrm{n_i}}, s) = \text{exp}(v_{s}^{\mathrm{o}\top}v_{s}^{\mathrm{n_i}}).
\end{gather}

\subsection{Definition Guided Fine-tuning}
\label{Method_DCL}
As described in Section~\ref{sec:Intro}, the meanings of words in a structured dictionary are very useful for human spell checking when spelling errors cannot be corrected with only phonetics and vision information.
To better utilize definition knowledge, we specially design the \emph{Definition Guided Fine-tuning} to make the model better understand the word meanings. Benefiting from the enhancement of definition knowledge, our model will be human-like, seeing spelling errors and associating them with their definitions, and then making reasonable corrections based on the original word meanings.

\paragraph{Positive and Negative Pairs}
\begin{CJK*}{UTF8}{gbsn}
As shown in Figure~\ref{Pos_Neg_Figure}, given a random training sample $\boldsymbol{x}_{\text{D}}^{\mathrm{o}}$ “举办一个误会” and its ground truth sentence $\boldsymbol{x}_{\text{D}}^{\mathrm{g}}$ “举办一个舞会”. To get the word meaning, we must first get the original word that contains the wrong position $s$. Therefore, we tokenize\footnote{We utilize the \href{https://github.com/hankcs/HanLP}{HanLP} to tokenize sentences into words.} the $\boldsymbol{x}_{\text{D}}^{\mathrm{g}}$ into words “举办/一个/舞会” and index the original word (i.e., “舞会”) in the dictionary\footnote{The pre-defined dictionary file we use is in the attachment.} to get its corresponding definition sentence as a positive sample $\boldsymbol{x}_{\text{D}}^{\mathrm{p}}$. As for the negative samples $\{\boldsymbol{x}_{\text{D}}^{\mathrm{n_i}}\}$, we will randomly select $N$ definition sentences of other words. 
\end{CJK*}

Considering that some words have multiple definitions, we design different word definition selection strategies as follows:
\begin{enumerate}
    \item \textbf{Select a random definition: }This is the easiest way to randomly select one sentence from multiple definition sentences.
    \item \textbf{Select the first definition: }Through preliminary analysis of the dictionary, we find that when a word has multiple definitions, the more forwardly positioned definition is often the more commonly used meaning of the word. Based on this observation, we propose to select the first definition to be the word meaning.
    \item \textbf{Select the most similar definition: }Intuitively, the meaning of a word can be revealed through its context. Therefore, we can also judge which definition sentence should be selected by the similarity between the sentence $\boldsymbol{x}_{\text{D}}^{\mathrm{g}}$ and the definition sentence. More practically, we obtain sentence representations through an encoder such as BERT~\cite{devlin-etal-2019-bert}, and further use the distance metric such as the cosine function to calculate the similarity between sentence representations.
\end{enumerate}
The effects of different word definition selection strategies will be analyzed in Section~\ref{Strategy_Experiments}.

\paragraph{Representation Metric}
When we tokenize the $\boldsymbol{x}_{\text{D}}^{\mathrm{g}}$, we obtain the index position of the original word in the sentence at the same time. Thus, assuming that the index positions of the original word are $[s, ..., s+w], s+w \leq T$, then we calculate the distance between representations as follows:
\begin{gather}
    f_{\text{D}}(\boldsymbol{d}^{\mathrm{o}}, \boldsymbol{d}^{\mathrm{p}}, s) = \text{cos}(\text{avg}([d_s^{\mathrm{o}}, ..., d_{s+w}^{\mathrm{o}}]), \text{avg}(\boldsymbol{d}^{\mathrm{p}})), \\
    f_{\text{D}}(\boldsymbol{d}^{\mathrm{o}}, \boldsymbol{d}^{\mathrm{n_i}}, s) = \text{cos}(\text{avg}([d_s^{\mathrm{o}}, ..., d_{s+w}^{\mathrm{o}}]), \text{avg}(\boldsymbol{d}^{\mathrm{n_i}})), 
\end{gather}
where the $\text{cos}(y_1, y_2)$ is the cosine distance, and the $\text{avg}([r_n, ..., r_m])$ is the mean pooling operation that calculates the average value of $[r_n, ..., r_m]$. In other words, the $\text{avg}([d_s^{\mathrm{o}}, ..., d_{s+w}^{\mathrm{o}}])$ is the representation of the phrase at index positions $[s, ..., s+w]$ in the sentence $\boldsymbol{x}_{\text{D}}^{\mathrm{o}}$ and the $\text{avg}(\boldsymbol{d}^{\mathrm{p}}), \{\text{avg}(\boldsymbol{d}^{\mathrm{n_i}})\}$ are the sentence representations of $\boldsymbol{x}_{\text{D}}^{\mathrm{p}},\{\boldsymbol{x}_{\text{D}}^{\mathrm{n_i}} \}$.

\subsection{Summary of Methodology}
In the above Sections~\ref{Method_PCL}-\ref{Method_DCL}, we describe in detail the contrastive learning objectives designed for the three types of knowledge. \emph{The purpose of these three kinds of contrastive learning objectives is to let the CSC model learn the external knowledge of phonetics, vision, and definition, and finally improve the model's CSC performance.} Additionally, because the model is to be used for the CSC task, it is still necessary to train the CSC training objective $\mathcal{L}_{\text{CSC}}$ with the CSC training data. So finally we have the following training loss:
\begin{equation}
    \mathcal{L} = \lambda_1\mathcal{L}_{\text{CSC}} + \lambda_2\mathcal{L}_{\text{P}} + \lambda_3\mathcal{L}_{\text{V}} + \lambda_4\mathcal{L}_{\text{D}}  ,
\end{equation}
where $\lambda_i$ is the task weighting. The $\mathcal{L}_{\text{CSC}}$ is the traditional CSC objective and the $\mathcal{L}_{\text{P}}, \mathcal{L}_{\text{V}}, \mathcal{L}_{\text{D}}$ are the contrastive objectives we design for “Phonetics, Vision, Definition” knowledge respectively.

\section{Experiments}
In this section, we first introduce the experiment settings and the main performance of \MethodName{}. Then we conduct detailed discussions and analyses to verify the effectiveness of our proposed methods.

\subsection{Datasets}
\paragraph{Training Data} In all our experiments, we use the widely used training data of most previous works~\cite{zhang-etal-2020-spelling,liu-etal-2021-plome,xu-etal-2021-read}, including the training sentences from SIGHAN13~\cite{wu-etal-2013-chinese}, SIGHAN14~\cite{yu-etal-2014-overview}, SIGHAN15~\cite{tseng-etal-2015-introduction}, and the generated training sentences (the size of this part data is 271K, we denote them as Wang271K in our paper)~\cite{wang-etal-2018-hybrid}. 

\paragraph{Test Data} To ensure the fairness of our experiments, we use the exact same test data as the baseline methods, which are from the SIGHAN13/14/15 test datasets. 
The details of the training/test data we use in our experiments are presented in Appendix~\ref{sec:dataset_appendix}.

\begin{table*}[h]
\small
\centering
\begin{tabular}{@{}c|l|p{0.90cm}p{0.90cm}p{0.90cm}|p{0.90cm}p{0.90cm}p{0.90cm}@{}}
\toprule
\multirow{2}{*}{Dataset} & \multicolumn{1}{c|}{\multirow{2}{*}{Method}} & \multicolumn{3}{c|}{Detection Level} & \multicolumn{3}{c}{Correction Level} \\
 & \multicolumn{1}{c|}{}  & Pre & Rec & F1  & Pre & Rec & F1 \\ \midrule
 
\multicolumn{1}{l|}{\multirow{9}{*}{SIGHAN13}} & SpellGCN~\cite{cheng-etal-2020-spellgcn} &  80.1 & 74.4 & 77.2 &  78.3 & 72.7 & 75.4 \\
\multicolumn{1}{l|}{} & MLM-phonetics~\cite{zhang-etal-2021-correcting} &  82.0 & 78.3 & 80.1 &  79.5 & 77.0 & 78.2 \\
\multicolumn{1}{l|}{} & DCN~\cite{wang-etal-2021-dynamic} &  86.8 & 79.6 & 83.0 &  84.7 & 77.7 & 81.0 \\
\multicolumn{1}{l|}{} & GAD~\cite{guo-etal-2021-global} &  85.7 & 79.5 & 82.5 &  84.9 & 78.7 & 81.6 \\
\multicolumn{1}{l|}{} & REALISE~\cite{xu-etal-2021-read} &  \textbf{\underline{88.6}} & \underline{82.5} & \underline{85.4} &  \textbf{\underline{87.2}} & \underline{81.2} & 84.1 \\ 
\multicolumn{1}{l|}{} & Two-Ways~\cite{li-etal-2021-exploration} &  - & - & 84.9 &  - & - & \underline{84.4} \\ 
\cmidrule(l){2-8} 
\multicolumn{1}{l|}{} & BERT~\cite{xu-etal-2021-read} &  85.0 & 77.0 & 80.8 &  83.0 & 75.2 & 78.9 \\
\multicolumn{1}{l|}{} & \MethodName{}  & 88.3 & \textbf{83.4} & \textbf{85.8} & \textbf{87.2} & \textbf{82.4} & \textbf{84.7} \\

\midrule

\multicolumn{1}{l|}{\multirow{9}{*}{SIGHAN14}} & SpellGCN~\cite{cheng-etal-2020-spellgcn}  & 65.1 & 69.5 & 67.2  & 63.1 & 67.2 & 65.3 \\
\multicolumn{1}{l|}{} & DCN~\cite{wang-etal-2021-dynamic}  & 67.4 & 70.4 & 68.9  & 65.8 & 68.7 & 67.2 \\
\multicolumn{1}{l|}{} & GAD~\cite{guo-etal-2021-global}  & 66.6 & 71.8 & 69.1  & 65.0 & 70.1 & 67.5 \\
\multicolumn{1}{l|}{} & REALISE~\cite{xu-etal-2021-read}  & \underline{67.8} & 71.5 & 69.6  & \underline{66.3} & 70.0 & 68.1 \\ 
\multicolumn{1}{l|}{} & Two-Ways~\cite{li-etal-2021-exploration}  & - & - & \underline{70.4}  & - & - & 68.6 \\
\multicolumn{1}{l|}{} & MLM-phonetics~\cite{zhang-etal-2021-correcting}  & 66.2 & \underline{\textbf{73.8}} & 69.8  & 64.2 & \underline{\textbf{73.8}} & \underline{68.7} \\ 
\cmidrule(l){2-8} 
\multicolumn{1}{l|}{} & BERT~\cite{xu-etal-2021-read}  & 64.5 & 68.6 & 66.5  & 62.4 & 66.3 & 64.3 \\
\multicolumn{1}{l|}{} & \MethodName{}   & \textbf{70.7} & 71.0 & \textbf{70.8} & \textbf{69.3} & 69.6 & \textbf{69.5} \\

\midrule

\multicolumn{1}{l|}{} & GAD~\cite{guo-etal-2021-global}  & 75.6 & 80.4 & 77.9  & 73.2 & 77.8 & 75.4 \\
\multicolumn{1}{l|}{\multirow{9}{*}{SIGHAN15}} & SpellGCN~\cite{cheng-etal-2020-spellgcn}  & 74.8 & 80.7 & 77.7  & 72.1 & 77.7 & 75.9 \\
\multicolumn{1}{l|}{} & DCN~\cite{wang-etal-2021-dynamic}  & 77.1 & 80.9 & 79.0  & 74.5 & 78.2 & 76.3 \\
\multicolumn{1}{l|}{} & PLOME~\cite{liu-etal-2021-plome}  & 77.4 & 81.5 & 79.4  & 75.3 & 79.3 & 77.2 \\
\multicolumn{1}{l|}{} & MLM-phonetics~\cite{zhang-etal-2021-correcting}  & \underline{77.5} & \underline{\textbf{83.1}} & \underline{80.2}  & 74.9 & \underline{80.2} & 77.5 \\ 
\multicolumn{1}{l|}{} & REALISE~\cite{xu-etal-2021-read}  & 77.3 & 81.3 & 79.3  & \underline{75.9} & 79.9 & 77.8 \\ 
\multicolumn{1}{l|}{} & Two-Ways~\cite{li-etal-2021-exploration}  & - & - & 80.0  & - & - & \underline{78.2} \\ 
\cmidrule(l){2-8} 
\multicolumn{1}{l|}{} & BERT~\cite{xu-etal-2021-read}  & 74.2 & 78.0 & 76.1  & 71.6 & 75.3 & 73.4 \\
\multicolumn{1}{l|}{} & \MethodName{}   & \textbf{79.2} & 82.8 & \textbf{80.9} & \textbf{77.6} & \textbf{81.2} & \textbf{79.3} \\

\bottomrule
\end{tabular}

\caption{The performance of \MethodName{} and baselines. For each dataset, we rank baselines from low to high performance according to the most important metric (i.e., correction level F1 score). Note that all results of baselines are directly from published papers. We \underline{underline} the previous state-of-the-art performance for convenient comparison. }
\label{Main_Results}
\end{table*}

\subsection{Baseline Methods}
To evaluate the performance of \MethodName{}, we select several latest CSC models as our baselines, including the previous state-of-the-art methods on SIGHAN13/14/15 datasets: \textbf{BERT}~\cite{devlin-etal-2019-bert} is fine-tuned on the training data only with the cross-entropy. \textbf{SpellGCN}~\cite{cheng-etal-2020-spellgcn} introduces the confusion set information through GCNs.
\textbf{GAD}~\cite{guo-etal-2021-global} combines a global attention decoder with BERT and trains the model under a confusion set guided replacement strategy.
\textbf{Two-Ways}~\cite{li-etal-2021-exploration} continually identifies the model's weak spots to generate more valuable training sentences. 
\textbf{DCN}~\cite{wang-etal-2021-dynamic} utilizes the Pinyin enhanced candidate generator and proposes the dynamic connected networks to build the dependencies. 
\textbf{MLM-phonetics}~\cite{zhang-etal-2021-correcting} introduces the phonetic features into the ERNIE~\cite{Sun_Wang_Li_Feng_Tian_Wu_Wang_2020} and uses the enhanced ERNIE model for CSC. \textbf{PLOME}~\cite{liu-etal-2021-plome} pre-trains BERT with a confusion set-based masking strategy and utilizes GRU~\cite{dey2017gate} to encode phonetics/strokes as input. 
\textbf{REALISE}~\cite{xu-etal-2021-read} is a multimodal model which mixes the semantic, phonetic, and graphic information to improve the model performance.

\subsection{Experimental Setup}
The character/sentence-level metrics are both used in the CSC task. According to the sentence-level metric, one test sentence will be judged to be correct only when all the wrong characters in it are detected and corrected successfully. Therefore, the sentence-level metric is stricter than the character-level metric because some sentences may have multiple wrong characters. So we report the sentence-level metrics for the evaluation in all our experiments, this setting is also widely used in previous works~\cite{li-etal-2021-exploration, liu-etal-2021-plome, xu-etal-2021-read}. More specifically, we report the metrics including Precision, Recall, and F1 score for detection and correction levels. At the detection level, all positions of wrong characters in a test sample should be detected correctly. At the correction level,  we require the model must not only detect but also correct all the spelling errors. Additionally, other implementation details of our experiments are shown in Appendix~\ref{sec:implementation_appendix}.

\subsection{Main Results}
From Table~\ref{Main_Results}, we observe that:
\begin{enumerate}
    \item Because \MethodName{} is essentially a fine-tuning framework of BERT, its direct baseline should be the BERT. The comparison results between \MethodName{} and BERT show that \MethodName{} outperforms BERT significantly on SIGHAN13/14/15, which verifies the effectiveness of our proposed heterogeneous knowledge guided fine-tuning methods.
    \item Compared with previous state-of-the-art models (i.e., Two-Ways, REALISE, and MLM-phonetics), our model utilizes only a thin BERT as the main body to achieve better performance, while REALISE and MLM-phonetics both explicitly introduce multimodal information into their inference process, which demonstrates the competitive performance of our proposed methods.
    \item Considering the effect of different knowledge, \MethodName{} is trained under the guidance of phonetics, vision, and definition knowledge, while most baselines (e.g., SpellGCN, DCN, and PLOME) also use different mechanisms to leverage the phonetics and vision knowledge. That our method outperforms these baselines indicates that the unique definition knowledge we focus on is very important for CSC.
\end{enumerate}

\subsection{Ablation Study}
We explore the effectiveness of each contrastive learning objective in \MethodName{} by conducting ablation studies with different variants. Specifically, in Table~\ref{tab:ablation}, $\text{MODEL}$ + $\text{K}$, $\text{K}\in\{\text{P, V, D}\}$  means that we use the CSC training objective $ \mathcal{L}_{\text{CSC}}$ and corresponding contrastive training objective $ \mathcal{L}_{\text{K}}$ to train the $\text{MODEL}$. Besides, because REALISE has its own way of using vision/phonetics features, which makes $\mathcal{L}_{\text{V}}$ and $ \mathcal{L}_{\text{P}}$ not meaningful, so we only perform $ \mathcal{L}_{\text{D}}$ on REALISE.

From the three rows of results using a single training objective (i.e., BERT+V/P/D), we know that each of our proposed contrastive learning strategies leads to significant performance improvements when applied to BERT alone. Particularly, the phenomenon that BERT+P outperforms BERT+V at the correction level is in line with the fact that 83\% of errors belong to phonological errors and 48\% belong to visual errors in the real scene~\cite{liu-etal-2021-plome}. Furthermore, we also see that all methods including the previous state-of-the-art model (i.e., REALISE) have further improvements after adding our proposed definition guided fine-tuning objective, which demonstrates that the definition information we focus on is very useful for enhancing CSC models.

\begin{table}[t]
\small
\centering

\begin{tabular}{@{}l|cc@{}}
\toprule
\multicolumn{1}{c|}{Method}  & Det-F1 & Cor-F1 \\
\midrule
~~BERT   & 76.1 & 73.4 \\
\quad + V(ision)  & 78.4 & 77.1 \\
\quad + P(honetics)  & 78.2 & 77.3 \\
\quad + D(efinition)  & 79.0 & 77.4 \\
\quad + V(ision)~ + P(honetics) & 79.6 & 78.1 \\ 
\quad + V(ision)~ + D(efinition) & 78.9 & 78.2 \\
\quad + P(honetics)~ + D(efinition) & 80.3 & 78.3 \\
\midrule
~~REALISE & 79.3 & 77.8 \\
\quad + D(efinition) & 80.3 & 78.6 \\
\midrule
~~\MethodName{}  & \textbf{80.9} & \textbf{79.3}  \\

\bottomrule
\end{tabular}

\caption{Ablation results on the SIGHAN15 test set. Note that the \MethodName{} is equivalent to BERT+V+P+D.}
\label{tab:ablation}
\end{table}

\subsection{Analysis and Discussion}

\begin{CJK*}{UTF8}{gbsn}
\begin{figure}[tp]
\centering
\subfigure[BERT (\textcolor{jiColor}{ji}/\textcolor{zhiColor}{zhi})] 
{ \label{bert-p} 
\includegraphics[height = 0.45 \columnwidth,width=0.45\columnwidth]{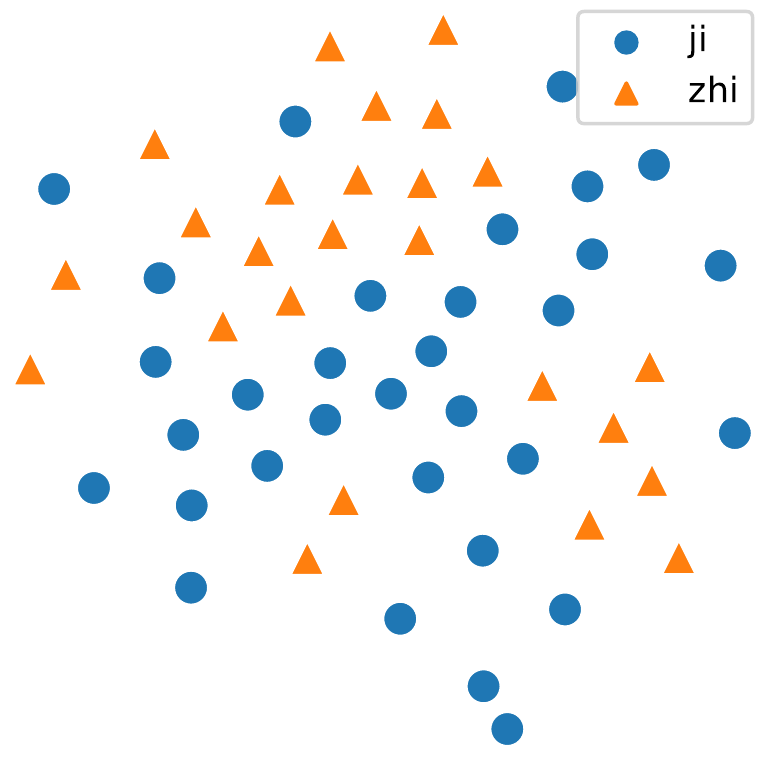} 
} 
\subfigure[BERT+P (\textcolor{jiColor}{ji}/\textcolor{zhiColor}{zhi})] 
{ \label{our-p} 
\includegraphics[height = 0.45 \columnwidth, width=0.45\columnwidth]{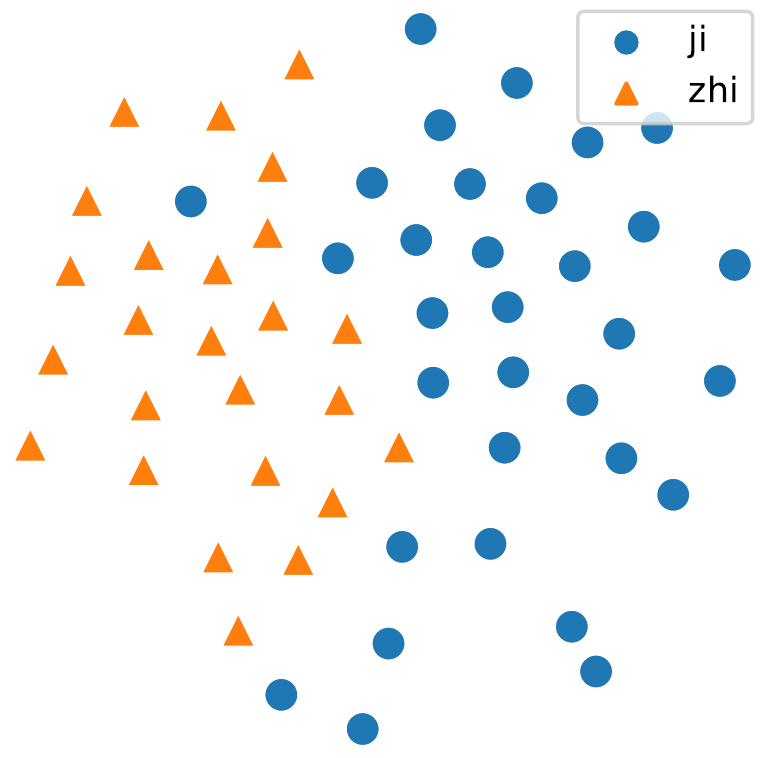} 
} 
\subfigure[BERT (\textcolor{xinColor}{新}/\textcolor{yingColor}{营})] 
{ \label{bert-v} 
\includegraphics[height = 0.45 \columnwidth, width=0.45\columnwidth]{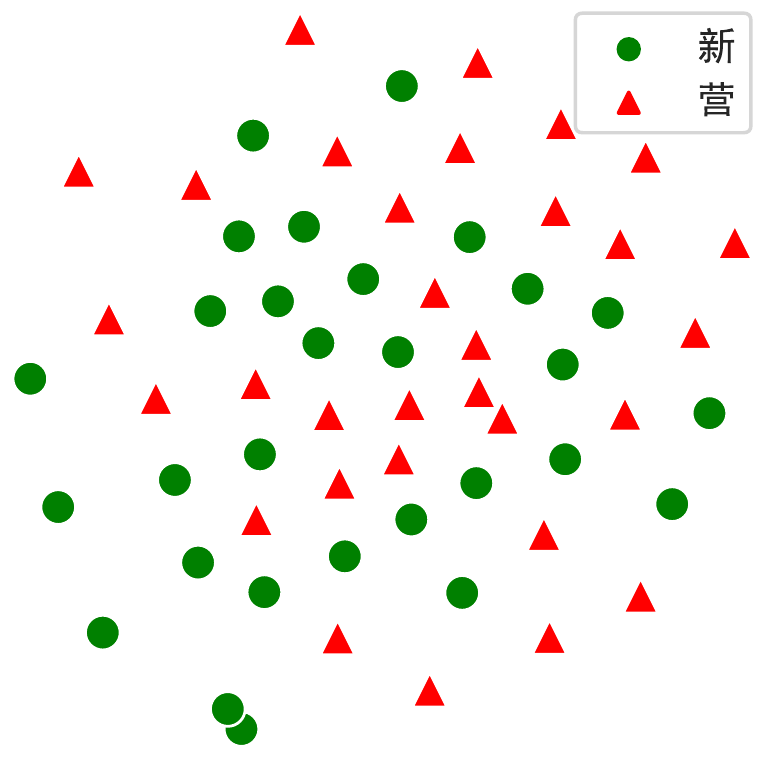} 
}
\subfigure[BERT+V (\textcolor{xinColor}{新}/\textcolor{yingColor}{营})] 
{ \label{our-v} 
\includegraphics[height = 0.45 \columnwidth, width=0.45\columnwidth]{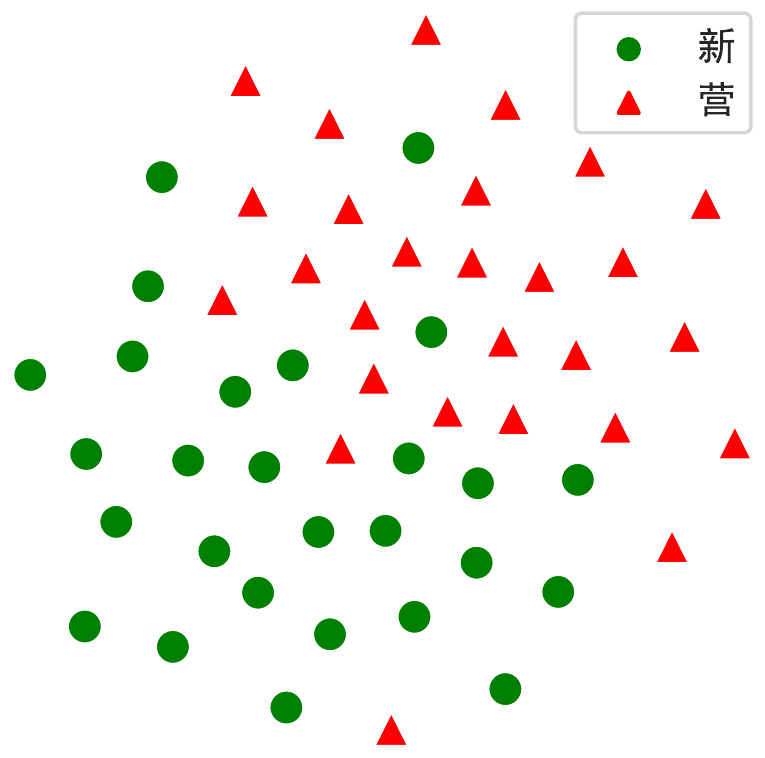} 
} 

\caption{Visualization (t-SNE) of phonetically/ visually similar characters. } 
\label{visualization} 
\end{figure}

\subsubsection{Visualization of Better Phonetic/Vision Representations}
The key motivation of our proposed phonetics/vision guided fine-tuning is to refine the representations of the models for characters on different dimensions of knowledge.
We hope that through the phonetics/vision guided fine-tuning, the model can be guided to represent characters with similar Pinyin/strokes closer, and characters with different Pinyin/strokes to represent farther.
Therefore, it is necessary to visualize the representations of the characters before and after the model is combined with our methods.
Specifically, we randomly select two groups of phonetically/visually similar characters (e.g., characters with similar Pinyin to “ji/zhi” and similar strokes to “新/营”), then apply BERT and BERT+P/V to obtain their representations. Finally, we use t-SNE to visualize these high-dimensional representations of characters.

Figure~\ref{visualization} shows the representation distribution of BERT and BERT+P/V for phonetically/visually similar characters. From the comparison of Figures~\ref{bert-p} and~\ref{our-p}, Figure~\ref{bert-p}’s representation of characters is messy, while in~\ref{our-p}, it can even be seen that there is a clear boundary between the two kinds of characters, which indicates that after the optimization of phonetics guided fine-tuning, it does represent the phonetically similar characters closer. Also in the visual comparison, we see that the points of the two colors in Figure~\ref{bert-v} are significantly more scattered, while Figure~\ref{our-v} is more orderly, which also verifies our motivation for proposing vision guided fine-tuning.
\end{CJK*}

\subsubsection{Effects of Different Word Definition Selection Strategies}
\label{Strategy_Experiments}

\begin{table}[t]
\small
\centering

\begin{tabular}{@{}l|p{0.90cm}p{0.90cm}p{0.90cm}@{}}
\toprule
\multicolumn{1}{c|}{Method} & Pre & Rec & F1 \\
\midrule
& \multicolumn{3}{c}{Detection Level} \\
\midrule
~~BERT  & 74.2 & 78.0 & 76.1 \\
~~\MethodName{}~(Random) & 77.7 & 81.3 & 79.5  \\
~~\MethodName{}~(First)  & 77.4 & 82.3 & 79.8 \\
~~\MethodName{}~(Similar)  & \textbf{79.2}  & \textbf{82.8}  & \textbf{80.9}  \\

\midrule
& \multicolumn{3}{c}{Correction Level} \\
\midrule
~~BERT  & 71.6 & 75.3 & 73.4 \\
~~\MethodName{}~(Random) & 75.8 & 80.6 & 78.1 \\
~~\MethodName{}~(First) & 76.7 & 80.2 & 78.4 \\
~~\MethodName{}~(Similar)   & \textbf{77.6} & \textbf{81.2} & \textbf{79.3} \\
\bottomrule
\end{tabular}

\caption{The results of \MethodName{} on SIGHAN15 when using different word definition selection strategies. }
\label{tab:strategy}
\end{table}

As mentioned in Section~\ref{Method_DCL}, we design three different word definition selection strategies for the definition guided fine-tuning, namely “select a random definition” (Random), “select the first definition” (First), and “select the most similar definition” (Similar). To further empirically explain why these strategies we proposed are effective, we conduct analyses as shown in Table~\ref{tab:strategy}. We apply \MethodName{} with different strategies on the SIGHAN15 dataset and observe the performance change.

From Table~\ref{tab:strategy}, we know that \MethodName{}~(Similar) has the best performance, followed by \MethodName{}~(First), and \MethodName{}~(Random) has the lowest improvement. Such results are consistent with the mechanism of these strategies. The better performance of \MethodName{}~(First) than \MethodName{}~(Random) shows that our observation on the dictionary is correct, that is, the first of multiple definitions of a word is often the most representative in most cases.
Additionally, the best performance of \MethodName{}~(Similar) also proves the effectiveness of our designed selection strategy that is based on sentence similarity.
It is worth mentioning that although the three strategies have different effects on the model performance, they all bring steady performance improvements compared to the baseline method (i.e., BERT).

\subsection{Case Study}

\begin{CJK*}{UTF8}{gbsn}

\begin{table}
\small
\centering
\begin{tabular}{p{1.4cm}p{5.2cm}}
\toprule

\textbf{Input 1:} & 要\textcolor{red}{永(\textit{\pinyin{yong3}})}于面对逆境。 \\
& Please \textcolor{red}{always} face adversity. \\ 
\midrule
\textbf{Output 1:} & 要\textcolor{blue}{勇(\textit{\pinyin{yong3}})}于面对逆境。 \\
 & Please face adversity \textcolor{blue}{bravely}.  \\ \midrule
  
\textbf{Input 2:} & 秋天\textcolor{red}{己}经无声的来到了。 \\
& Autumn \textcolor{red}{self} come silently. \\ 
\midrule
\textbf{Output 2:} & 秋天\textcolor{blue}{已}经无声的来到了。  \\
 & Autumn \textcolor{blue}{has} come silently. \\ \midrule

\textbf{Input 3:} & 迎接每一个\textcolor{red}{固}难，并\textcolor{orange}{克服}它。 \\
& Meet every \textcolor{red}{hardship} and \textcolor{orange}{overcome} it. \\ 
\midrule
\textbf{Output 3:} & 迎接每一个\textcolor{blue}{困}难，并\textcolor{orange}{克服}它。  \\
 & Meet every \textcolor{blue}{difficulty} and \textcolor{orange}{overcome} it. \\ 
\textbf{Definition:} & 【\textcolor{blue}{困}难】: (名) 工作、生活中遇到的不易解决的问题或障碍, \textcolor{orange}{克服}～ \\
& (noun) Problems or obstacles in work and life that are not easy to solve, \textcolor{orange}{overcome}～\\
\bottomrule
\end{tabular}
\caption{Examples of the input/output of \MethodName{}. 
We mark the \textcolor{red}{wrong}/\textcolor{blue}{correct}/\textcolor{orange}{key} characters.  }
\label{Case_Studies}
\end{table}

From the first/second cases in Table~\ref{Case_Studies}, we know that our \MethodName{} perceives the phonetic and visual similarity of Chinese characters, so as to accurately detect the wrong positions and make reasonable corrections. Particularly, for the first example, if ignoring the phonetic similarity, there are other candidate characters such as “乐(\textit{\pinyin{le4}})” and “敢(\textit{\pinyin{yong3}})”. But the \MethodName{}'s output is the best correction because it is more in line with the essential of CSC. Additionally, in the third example, “固(\textit{\pinyin{gu4}})” and “困(\textit{\pinyin{kun4}})” are neither phonetically nor visually similar, and \MethodName{} successfully correct this case because it perceives the definition of “困难” in the dictionary. Without the help of the definition, we can replace the “固(\textit{\pinyin{gu4}})” with the “苦(\textit{\pinyin{ku3}})” which is more phonetically similar to “固(\textit{\pinyin{gu4}})”. However, in daily use of Chinese, the combination of “克服” and “苦难” is not common. Therefore, this example just reflects the importance of definition knowledge we are concerned with for CSC.

\end{CJK*}

\section{Conclusion}
In this paper, we propose to promote CSC by utilizing various knowledge contained in the dictionary.
We introduce \MethodName{}, a unified fine-tuning framework that aims to perform contrastive learning for three kinds of heterogeneous knowledge.
Extensive experiments and empirical analyses verify the motivation of our study and the effectiveness of our proposed methods.
The dictionary knowledge we focus on is not only beneficial for CSC, but also crucial for other Chinese text processing tasks. Therefore, in the future, we will continue to mine the knowledge contained in the dictionary to improve other Chinese text processing tasks.

\section{Limitations}
In this section, we discuss the limitations of our work in detail and propose corresponding solutions that we believe are feasible.
\subsection{Language Limitation}
Our work and the proposed method focus on the Chinese Spell Checking (CSC) task. The language characteristics of Chinese are very different from other languages such as English. For example, the phonetically or visually characters, which bring great challenge to CSC, does not exist in English. Therefore, the limitation of language characteristics prevents our method from being directly transferable to English scenarios. However, we also believe that the definition knowledge in the dictionary we are concerned with still has important implications for English text error correction.

\subsection{Encoder Selection}
Our proposed \MethodName{} framework is a unified fine-tuning framework to guide the CSC models to learn heterogeneous knowledge. The unified paradigm allows \MethodName{} to impose no strict restrictions on the various encoders used in it. To verify the effectiveness of \MethodName{}, in our experiments, we just choose the simple configuration as $\mathrm{E}_{\text{P}}$, $\mathrm{E}_{\text{V}}$, $\mathrm{E}_{\text{D}}$ (see Appendix~\ref{sec:implementation_appendix}). In the future, we suggest that more complex models and configurations can be used for more performance improvements.

\subsection{Running Efficiency}
As academic verification experiments, we do not consider the running efficiency of our proposed methods in the specific code implementation. Specifically, it takes about 10 hours on 1 V100 GPU to finish the training process and it takes up to 24G GPU memory. We think that there are at least two solutions to improve efficiency: (1) Deploying the model training process to multi-GPUs and using data-parallel operations can increase the training batch size and shorten the training time. (2) Change the online positive and negative sample construction to offline, that is, various positive and negative sample pairs for training are constructed and stored in advance, which can also greatly save the time cost during training.

\section*{Acknowledgement}
This research is supported by National Natural Science Foundation of China (Grant No.62276154 and 62011540405), Beijing Academy of Artificial Intelligence (BAAI), the Natural Science Foundation of Guangdong Province (Grant No. 2021A1515012640), Basic Research Fund of Shenzhen City (Grant No. JCYJ20210324120012033 and JCYJ20190813165003837), and Overseas Cooperation Research Fund of Tsinghua Shenzhen International Graduate School  (Grant No. HW2021008).

\bibliography{anthology,custom}

\begin{thebibliography}{38}
\expandafter\ifx\csname natexlab\endcsname\relax\def\natexlab#1{#1}\fi

\bibitem[{Chen et~al.(2020)Chen, Kornblith, Norouzi, and
  Hinton}]{chen2020simple}
Ting Chen, Simon Kornblith, Mohammad Norouzi, and Geoffrey Hinton. 2020.
\newblock A simple framework for contrastive learning of visual
  representations.
\newblock In \emph{International conference on machine learning}, pages
  1597--1607. PMLR.

\bibitem[{Cheng et~al.(2020)Cheng, Xu, Chen, Jiang, Wang, Wang, Chu, and
  Qi}]{cheng-etal-2020-spellgcn}
Xingyi Cheng, Weidi Xu, Kunlong Chen, Shaohua Jiang, Feng Wang, Taifeng Wang,
  Wei Chu, and Yuan Qi. 2020.
\newblock \href {https://doi.org/10.18653/v1/2020.acl-main.81} {{S}pell{GCN}:
  Incorporating phonological and visual similarities into language models for
  {C}hinese spelling check}.
\newblock In \emph{Proceedings of the 58th Annual Meeting of the Association
  for Computational Linguistics}, pages 871--881, Online. Association for
  Computational Linguistics.

\bibitem[{Cui et~al.(2020)Cui, Che, Liu, Qin, Wang, and
  Hu}]{cui-etal-2020-revisiting}
Yiming Cui, Wanxiang Che, Ting Liu, Bing Qin, Shijin Wang, and Guoping Hu.
  2020.
\newblock \href {https://www.aclweb.org/anthology/2020.findings-emnlp.58}
  {Revisiting pre-trained models for {C}hinese natural language processing}.
\newblock In \emph{Proceedings of the 2020 Conference on Empirical Methods in
  Natural Language Processing: Findings}, pages 657--668, Online. Association
  for Computational Linguistics.

\bibitem[{Devlin et~al.(2019)Devlin, Chang, Lee, and
  Toutanova}]{devlin-etal-2019-bert}
Jacob Devlin, Ming-Wei Chang, Kenton Lee, and Kristina Toutanova. 2019.
\newblock \href {https://doi.org/10.18653/v1/N19-1423} {{BERT}: Pre-training of
  deep bidirectional transformers for language understanding}.
\newblock In \emph{Proceedings of the 2019 Conference of the North {A}merican
  Chapter of the Association for Computational Linguistics: Human Language
  Technologies, Volume 1 (Long and Short Papers)}, pages 4171--4186,
  Minneapolis, Minnesota. Association for Computational Linguistics.

\bibitem[{Dey and Salem(2017)}]{dey2017gate}
Rahul Dey and Fathi~M Salem. 2017.
\newblock Gate-variants of gated recurrent unit (gru) neural networks.
\newblock In \emph{2017 IEEE 60th international midwest symposium on circuits
  and systems (MWSCAS)}, pages 1597--1600. IEEE.

\bibitem[{Gao et~al.(2021)Gao, Yao, and Chen}]{gao2021simcse}
Tianyu Gao, Xingcheng Yao, and Danqi Chen. 2021.
\newblock Simcse: Simple contrastive learning of sentence embeddings.
\newblock \emph{arXiv preprint arXiv:2104.08821}.

\bibitem[{Guo et~al.(2021)Guo, Ni, Wang, Zhu, and Xie}]{guo-etal-2021-global}
Zhao Guo, Yuan Ni, Keqiang Wang, Wei Zhu, and Guotong Xie. 2021.
\newblock \href {https://doi.org/10.18653/v1/2021.findings-acl.122} {Global
  attention decoder for {C}hinese spelling error correction}.
\newblock In \emph{Findings of the Association for Computational Linguistics:
  ACL-IJCNLP 2021}, pages 1419--1428, Online. Association for Computational
  Linguistics.

\bibitem[{Hadsell et~al.(2006)Hadsell, Chopra, and LeCun}]{HadsellCon}
R.~Hadsell, S.~Chopra, and Y.~LeCun. 2006.
\newblock \href {https://doi.org/10.1109/CVPR.2006.100} {Dimensionality
  reduction by learning an invariant mapping}.
\newblock In \emph{2006 IEEE Computer Society Conference on Computer Vision and
  Pattern Recognition (CVPR'06)}, volume~2, pages 1735--1742.

\bibitem[{He et~al.(2020{\natexlab{a}})He, Fan, Wu, Xie, and
  Girshick}]{he2020momentum}
Kaiming He, Haoqi Fan, Yuxin Wu, Saining Xie, and Ross Girshick.
  2020{\natexlab{a}}.
\newblock Momentum contrast for unsupervised visual representation learning.
\newblock In \emph{Proceedings of the IEEE/CVF conference on computer vision
  and pattern recognition}, pages 9729--9738.

\bibitem[{He et~al.(2020{\natexlab{b}})He, Fan, Wu, Xie, and
  Girshick}]{DBLP:conf/cvpr/He0WXG20}
Kaiming He, Haoqi Fan, Yuxin Wu, Saining Xie, and Ross~B. Girshick.
  2020{\natexlab{b}}.
\newblock \href {https://doi.org/10.1109/CVPR42600.2020.00975} {Momentum
  contrast for unsupervised visual representation learning}.
\newblock In \emph{2020 {IEEE/CVF} Conference on Computer Vision and Pattern
  Recognition, {CVPR} 2020, Seattle, WA, USA, June 13-19, 2020}, pages
  9726--9735. Computer Vision Foundation / {IEEE}.

\bibitem[{Hong et~al.(2019)Hong, Yu, He, Liu, and Liu}]{hong-etal-2019-faspell}
Yuzhong Hong, Xianguo Yu, Neng He, Nan Liu, and Junhui Liu. 2019.
\newblock \href {https://doi.org/10.18653/v1/D19-5522} {{FASP}ell: A fast,
  adaptable, simple, powerful {C}hinese spell checker based on {DAE}-decoder
  paradigm}.
\newblock In \emph{Proceedings of the 5th Workshop on Noisy User-generated Text
  (W-NUT 2019)}, pages 160--169, Hong Kong, China. Association for
  Computational Linguistics.

\bibitem[{Huang et~al.(2021)Huang, Li, Jiang, Zhang, Chen, Wang, and
  Xiao}]{huang-etal-2021-phmospell}
Li~Huang, Junjie Li, Weiwei Jiang, Zhiyu Zhang, Minchuan Chen, Shaojun Wang,
  and Jing Xiao. 2021.
\newblock \href {https://doi.org/10.18653/v1/2021.acl-long.464} {{PHMOS}pell:
  Phonological and morphological knowledge guided {C}hinese spelling check}.
\newblock In \emph{Proceedings of the 59th Annual Meeting of the Association
  for Computational Linguistics and the 11th International Joint Conference on
  Natural Language Processing (Volume 1: Long Papers)}, pages 5958--5967,
  Online. Association for Computational Linguistics.

\bibitem[{Ji et~al.(2021)Ji, Yan, and Qiu}]{ji-etal-2021-spellbert}
Tuo Ji, Hang Yan, and Xipeng Qiu. 2021.
\newblock \href {https://doi.org/10.18653/v1/2021.emnlp-main.287}
  {{S}pell{BERT}: A lightweight pretrained model for {C}hinese spelling check}.
\newblock In \emph{Proceedings of the 2021 Conference on Empirical Methods in
  Natural Language Processing}, pages 3544--3551, Online and Punta Cana,
  Dominican Republic. Association for Computational Linguistics.

\bibitem[{Khosla et~al.(2020)Khosla, Teterwak, Wang, Sarna, Tian, Isola,
  Maschinot, Liu, and Krishnan}]{khosla2020supervised}
Prannay Khosla, Piotr Teterwak, Chen Wang, Aaron Sarna, Yonglong Tian, Phillip
  Isola, Aaron Maschinot, Ce~Liu, and Dilip Krishnan. 2020.
\newblock Supervised contrastive learning.
\newblock \emph{arXiv preprint arXiv:2004.11362}.

\bibitem[{Kim et~al.(2021)Kim, Yoo, and Lee}]{kim-etal-2021-self}
Taeuk Kim, Kang~Min Yoo, and Sang-goo Lee. 2021.
\newblock \href {https://doi.org/10.18653/v1/2021.acl-long.197} {Self-guided
  contrastive learning for {BERT} sentence representations}.
\newblock In \emph{Proceedings of the 59th Annual Meeting of the Association
  for Computational Linguistics and the 11th International Joint Conference on
  Natural Language Processing (Volume 1: Long Papers)}, pages 2528--2540,
  Online. Association for Computational Linguistics.

\bibitem[{Kipf and Welling(2017)}]{kipf2017semi}
Thomas~N. Kipf and Max Welling. 2017.
\newblock Semi-supervised classification with graph convolutional networks.
\newblock In \emph{International Conference on Learning Representations
  (ICLR)}.

\bibitem[{Li et~al.(2021)Li, Zhang, Zheng, and
  Huang}]{li-etal-2021-exploration}
Chong Li, Cenyuan Zhang, Xiaoqing Zheng, and Xuanjing Huang. 2021.
\newblock \href {https://doi.org/10.18653/v1/2021.acl-short.56} {Exploration
  and exploitation: Two ways to improve {C}hinese spelling correction models}.
\newblock In \emph{Proceedings of the 59th Annual Meeting of the Association
  for Computational Linguistics and the 11th International Joint Conference on
  Natural Language Processing (Volume 2: Short Papers)}, pages 441--446,
  Online. Association for Computational Linguistics.

\bibitem[{Li et~al.(2022{\natexlab{a}})Li, Li, He, Yu, Shen, and
  Zheng}]{li2022contrastive}
Yinghui Li, Yangning Li, Yuxin He, Tianyu Yu, Ying Shen, and Hai-Tao Zheng.
  2022{\natexlab{a}}.
\newblock Contrastive learning with hard negative entities for entity set
  expansion.
\newblock \emph{arXiv preprint arXiv:2204.07789}.

\bibitem[{Li et~al.(2022{\natexlab{b}})Li, Zhou, Li, Li, Liu, Sun, Wang, Li,
  Cao, and Zheng}]{li-etal-2022-past}
Yinghui Li, Qingyu Zhou, Yangning Li, Zhongli Li, Ruiyang Liu, Rongyi Sun,
  Zizhen Wang, Chao Li, Yunbo Cao, and Hai-Tao Zheng. 2022{\natexlab{b}}.
\newblock \href {https://aclanthology.org/2022.findings-acl.252} {The past
  mistake is the future wisdom: Error-driven contrastive probability
  optimization for {C}hinese spell checking}.
\newblock In \emph{Findings of the Association for Computational Linguistics:
  ACL 2022}, pages 3202--3213, Dublin, Ireland. Association for Computational
  Linguistics.

\bibitem[{Liu et~al.(2010)Liu, Lai, Chuang, and Lee}]{liu-etal-2010-visually}
Chao-Lin Liu, Min-Hua Lai, Yi-Hsuan Chuang, and Chia-Ying Lee. 2010.
\newblock \href {https://aclanthology.org/C10-2085} {Visually and
  phonologically similar characters in incorrect simplified {C}hinese words}.
\newblock In \emph{Coling 2010: Posters}, pages 739--747, Beijing, China.
  Coling 2010 Organizing Committee.

\bibitem[{Liu et~al.(2021)Liu, Yang, Yue, Zhang, and
  Wang}]{liu-etal-2021-plome}
Shulin Liu, Tao Yang, Tianchi Yue, Feng Zhang, and Di~Wang. 2021.
\newblock \href {https://doi.org/10.18653/v1/2021.acl-long.233} {{PLOME}:
  Pre-training with misspelled knowledge for {C}hinese spelling correction}.
\newblock In \emph{Proceedings of the 59th Annual Meeting of the Association
  for Computational Linguistics and the 11th International Joint Conference on
  Natural Language Processing (Volume 1: Long Papers)}, pages 2991--3000,
  Online. Association for Computational Linguistics.

\bibitem[{Loshchilov and Hutter(2018)}]{loshchilov2018fixing}
Ilya Loshchilov and Frank Hutter. 2018.
\newblock Fixing weight decay regularization in adam.

\bibitem[{Lyu et~al.(2021)Lyu, Chen, and Yu}]{lyu-etal-2021-glyph-enhanced}
Boer Lyu, Lu~Chen, and Kai Yu. 2021.
\newblock \href {https://doi.org/10.18653/v1/2021.findings-emnlp.386} {Glyph
  enhanced {C}hinese character pre-training for lexical sememe prediction}.
\newblock In \emph{Findings of the Association for Computational Linguistics:
  EMNLP 2021}, pages 4549--4555, Punta Cana, Dominican Republic. Association
  for Computational Linguistics.

\bibitem[{Paszke et~al.(2019)Paszke, Gross, Massa, Lerer, Bradbury, Chanan,
  Killeen, Lin, Gimelshein, Antiga et~al.}]{paszke2019pytorch}
Adam Paszke, Sam Gross, Francisco Massa, Adam Lerer, James Bradbury, Gregory
  Chanan, Trevor Killeen, Zeming Lin, Natalia Gimelshein, Luca Antiga, et~al.
  2019.
\newblock Pytorch: An imperative style, high-performance deep learning library.
\newblock \emph{Advances in neural information processing systems},
  32:8026--8037.

\bibitem[{Qin et~al.(2021)Qin, Lin, Takanobu, Liu, Li, Ji, Huang, Sun, and
  Zhou}]{qin-etal-2021-erica}
Yujia Qin, Yankai Lin, Ryuichi Takanobu, Zhiyuan Liu, Peng Li, Heng Ji, Minlie
  Huang, Maosong Sun, and Jie Zhou. 2021.
\newblock \href {https://doi.org/10.18653/v1/2021.acl-long.260} {{ERICA}:
  Improving entity and relation understanding for pre-trained language models
  via contrastive learning}.
\newblock In \emph{Proceedings of the 59th Annual Meeting of the Association
  for Computational Linguistics and the 11th International Joint Conference on
  Natural Language Processing (Volume 1: Long Papers)}, pages 3350--3363,
  Online. Association for Computational Linguistics.

\bibitem[{Sun et~al.(2020)Sun, Wang, Li, Feng, Tian, Wu, and
  Wang}]{Sun_Wang_Li_Feng_Tian_Wu_Wang_2020}
Yu~Sun, Shuohuan Wang, Yukun Li, Shikun Feng, Hao Tian, Hua Wu, and Haifeng
  Wang. 2020.
\newblock \href {https://doi.org/10.1609/aaai.v34i05.6428} {Ernie 2.0: A
  continual pre-training framework for language understanding}.
\newblock \emph{Proceedings of the AAAI Conference on Artificial Intelligence},
  34(05):8968--8975.

\bibitem[{Tseng et~al.(2015)Tseng, Lee, Chang, and
  Chen}]{tseng-etal-2015-introduction}
Yuen-Hsien Tseng, Lung-Hao Lee, Li-Ping Chang, and Hsin-Hsi Chen. 2015.
\newblock \href {https://doi.org/10.18653/v1/W15-3106} {Introduction to
  {SIGHAN} 2015 bake-off for {C}hinese spelling check}.
\newblock In \emph{Proceedings of the Eighth {SIGHAN} Workshop on {C}hinese
  Language Processing}, pages 32--37, Beijing, China. Association for
  Computational Linguistics.

\bibitem[{van~den Oord et~al.(2018)van~den Oord, Li, and
  Vinyals}]{DBLP:journals/corr/abs-1807-03748}
A{\"{a}}ron van~den Oord, Yazhe Li, and Oriol Vinyals. 2018.
\newblock \href {http://arxiv.org/abs/1807.03748} {Representation learning with
  contrastive predictive coding}.
\newblock \emph{CoRR}, abs/1807.03748.

\bibitem[{Wang et~al.(2021)Wang, Che, Wu, Wang, Hu, and
  Liu}]{wang-etal-2021-dynamic}
Baoxin Wang, Wanxiang Che, Dayong Wu, Shijin Wang, Guoping Hu, and Ting Liu.
  2021.
\newblock \href {https://doi.org/10.18653/v1/2021.findings-acl.216} {Dynamic
  connected networks for {C}hinese spelling check}.
\newblock In \emph{Findings of the Association for Computational Linguistics:
  ACL-IJCNLP 2021}, pages 2437--2446, Online. Association for Computational
  Linguistics.

\bibitem[{Wang et~al.(2018)Wang, Song, Li, Han, and
  Zhang}]{wang-etal-2018-hybrid}
Dingmin Wang, Yan Song, Jing Li, Jialong Han, and Haisong Zhang. 2018.
\newblock \href {https://doi.org/10.18653/v1/D18-1273} {A hybrid approach to
  automatic corpus generation for {C}hinese spelling check}.
\newblock In \emph{Proceedings of the 2018 Conference on Empirical Methods in
  Natural Language Processing}, pages 2517--2527, Brussels, Belgium.
  Association for Computational Linguistics.

\bibitem[{Wang et~al.(2019)Wang, Tay, and Zhong}]{wang-etal-2019-confusionset}
Dingmin Wang, Yi~Tay, and Li~Zhong. 2019.
\newblock \href {https://doi.org/10.18653/v1/P19-1578} {Confusionset-guided
  pointer networks for {C}hinese spelling check}.
\newblock In \emph{Proceedings of the 57th Annual Meeting of the Association
  for Computational Linguistics}, pages 5780--5785, Florence, Italy.
  Association for Computational Linguistics.

\bibitem[{Wolf et~al.(2020)Wolf, Debut, Sanh, Chaumond, Delangue, Moi, Cistac,
  Rault, Louf, Funtowicz, Davison, Shleifer, von Platen, Ma, Jernite, Plu, Xu,
  Le~Scao, Gugger, Drame, Lhoest, and Rush}]{wolf-etal-2020-transformers}
Thomas Wolf, Lysandre Debut, Victor Sanh, Julien Chaumond, Clement Delangue,
  Anthony Moi, Pierric Cistac, Tim Rault, Remi Louf, Morgan Funtowicz, Joe
  Davison, Sam Shleifer, Patrick von Platen, Clara Ma, Yacine Jernite, Julien
  Plu, Canwen Xu, Teven Le~Scao, Sylvain Gugger, Mariama Drame, Quentin Lhoest,
  and Alexander Rush. 2020.
\newblock \href {https://doi.org/10.18653/v1/2020.emnlp-demos.6} {Transformers:
  State-of-the-art natural language processing}.
\newblock In \emph{Proceedings of the 2020 Conference on Empirical Methods in
  Natural Language Processing: System Demonstrations}, pages 38--45, Online.
  Association for Computational Linguistics.

\bibitem[{Wu et~al.(2013{\natexlab{a}})Wu, Chiu, and
  Chang}]{wu-etal-2013-integrating}
Jian-cheng Wu, Hsun-wen Chiu, and Jason~S. Chang. 2013{\natexlab{a}}.
\newblock \href {https://aclanthology.org/O13-5002} {Integrating dictionary and
  web n-grams for {C}hinese spell checking}.
\newblock In \emph{International Journal of Computational Linguistics {\&}
  {C}hinese Language Processing, Volume 18, Number 4, {D}ecember 2013-Special
  Issue on Selected Papers from {ROCLING} {XXV}}.

\bibitem[{Wu et~al.(2013{\natexlab{b}})Wu, Liu, and Lee}]{wu-etal-2013-chinese}
Shih-Hung Wu, Chao-Lin Liu, and Lung-Hao Lee. 2013{\natexlab{b}}.
\newblock \href {https://aclanthology.org/W13-4406} {{C}hinese spelling check
  evaluation at {SIGHAN} bake-off 2013}.
\newblock In \emph{Proceedings of the Seventh {SIGHAN} Workshop on {C}hinese
  Language Processing}, pages 35--42, Nagoya, Japan. Asian Federation of
  Natural Language Processing.

\bibitem[{Xu et~al.(2021)Xu, Li, Zhou, Li, Wang, Cao, Huang, and
  Mao}]{xu-etal-2021-read}
Heng-Da Xu, Zhongli Li, Qingyu Zhou, Chao Li, Zizhen Wang, Yunbo Cao, Heyan
  Huang, and Xian-Ling Mao. 2021.
\newblock \href {https://doi.org/10.18653/v1/2021.findings-acl.64} {Read,
  listen, and see: Leveraging multimodal information helps {C}hinese spell
  checking}.
\newblock In \emph{Findings of the Association for Computational Linguistics:
  ACL-IJCNLP 2021}, pages 716--728, Online. Association for Computational
  Linguistics.

\bibitem[{Yu et~al.(2014)Yu, Lee, Tseng, and Chen}]{yu-etal-2014-overview}
Liang-Chih Yu, Lung-Hao Lee, Yuen-Hsien Tseng, and Hsin-Hsi Chen. 2014.
\newblock \href {https://doi.org/10.3115/v1/W14-6820} {Overview of {SIGHAN}
  2014 bake-off for {C}hinese spelling check}.
\newblock In \emph{Proceedings of The Third {CIPS}-{SIGHAN} Joint Conference on
  {C}hinese Language Processing}, pages 126--132, Wuhan, China. Association for
  Computational Linguistics.

\bibitem[{Zhang et~al.(2021)Zhang, Pang, Zhang, Wang, He, Sun, Wu, and
  Wang}]{zhang-etal-2021-correcting}
Ruiqing Zhang, Chao Pang, Chuanqiang Zhang, Shuohuan Wang, Zhongjun He, Yu~Sun,
  Hua Wu, and Haifeng Wang. 2021.
\newblock \href {https://doi.org/10.18653/v1/2021.findings-acl.198} {Correcting
  {C}hinese spelling errors with phonetic pre-training}.
\newblock In \emph{Findings of the Association for Computational Linguistics:
  ACL-IJCNLP 2021}, pages 2250--2261, Online. Association for Computational
  Linguistics.

\bibitem[{Zhang et~al.(2020)Zhang, Huang, Liu, and
  Li}]{zhang-etal-2020-spelling}
Shaohua Zhang, Haoran Huang, Jicong Liu, and Hang Li. 2020.
\newblock \href {https://doi.org/10.18653/v1/2020.acl-main.82} {Spelling error
  correction with soft-masked {BERT}}.
\newblock In \emph{Proceedings of the 58th Annual Meeting of the Association
  for Computational Linguistics}, pages 882--890, Online. Association for
  Computational Linguistics.

\end{thebibliography}
\bibliographystyle{acl_natbib}

\clearpage
\appendix

\section{Appendix}

\subsection{Datasets Details}
\label{sec:dataset_appendix}
Please kindly note that the original sentences of SIGHAN datasets are in Traditional Chinese, so we need to convert these original texts to Simplified Chinese using the OpenCC tool\footnote{https://github.com/BYVoid/OpenCC}. This data pre-process procedure has been widely used in previously published works~\cite{wang-etal-2019-confusionset,cheng-etal-2020-spellgcn,zhang-etal-2020-spelling}. The details of the datasets we use in our experiments are presented in Table~\ref{Data_Statistics}.

\begin{table}[h]
\small
\centering
\begin{tabular}{lrrr}
\hline Training Data & \#Sent & Avg. Length & \#Errors \\
\hline SIGHAN13 & 700 & $41.8$ & 343 \\
SIGHAN14 & 3,437 & $49.6$ & 5,122 \\
SIGHAN15 & 2,338 & $31.3$ & 3,037 \\
Wang271K & 271,329 & $42.6$ & 381,962 \\
\hline Total & 277,804 & $42.6$ & 390464 \\
\hline \hline Test Data & \#Sent & Avg. Length & \#Errors \\
\hline SIGHAN13 & 1,000 & $74.3$ & 1,224 \\
SIGHAN14 & 1,062 & $50.0$ & 771 \\
SIGHAN15 & 1,100 & $30.6$ & 703 \\
\hline Total & 3,162 & $50.9$ & 2,698 \\
\hline
\end{tabular}

\caption{Statistics of the datasets that we use in experiments. We report the number of sentences (\#Sent), the average sentence length (Avg.Length), and the number of spelling errors (\#Errors).}
\label{Data_Statistics}
\end{table}

\subsection{Implementation Details}
\label{sec:implementation_appendix}
In our experiments, all the source code is implemented using Pytorch~\cite{paszke2019pytorch} based on the Huggingface's Transformer library\footnote{https://github.com/huggingface/transformers}~\cite{wolf-etal-2020-transformers}. 
For the implementation of $\mathrm{E}_{\text{C}}$, we use the cross-entropy function as the $\mathcal{L}_{\text{CSC}}$ and BERT as the main CSC model.
The BERT's architecture we use in our experiments is the same as the~$BERT_{BASE}$, which has 12 transformers layers with 12 attention heads and its hidden state size is 768. And the initial weights of BERT are from the weights of Chinese BERT-wwm~\cite{cui-etal-2020-revisiting}. 
For the implementation of $\mathrm{E}_{\text{P}}$, $\mathrm{E}_{\text{V}}$, $\mathrm{E}_{\text{D}}$, we preliminarily select the BERT consistent with the above description as $\mathrm{E}_{\text{P}}$ and $\mathrm{E}_{\text{D}}$, and we use the glyph enhanced pre-training model proposed in~\citet{lyu-etal-2021-glyph-enhanced} as $\mathrm{E}_{\text{V}}$ to obtain the strokes representations of Chinese characters.

We set the maximum sentence length to 128.
We train \MethodName{} with the AdamW optimizer~\cite{loshchilov2018fixing} for 10 epochs and set the training batch size to 32. 
The model is trained with learning rate warming up and linear decay, while the initial learning rate is set to 5e-5.
The negative pairs size $N$ of a mini-batch is set to 8 when we report the main results of \MethodName{}.
Besides, the weighting factors~$\lambda_i$ of $\mathcal{L}$ are all set to 1. 

As mentioned in~\cite{cheng-etal-2020-spellgcn,xu-etal-2021-read, li-etal-2022-past}, lots of the mixed usage of auxiliary (such as “\begin{CJK*}{UTF8}{gbsn}的\end{CJK*}”, “\begin{CJK*}{UTF8}{gbsn}地\end{CJK*}”, and “\begin{CJK*}{UTF8}{gbsn}得\end{CJK*}”) are wrongly annotated, which makes the quality of the SIGHAN13 test dataset very poor. To alleviate this problem and more accurately evaluate the performance of models on SIGHAN13, there exist two main solutions in previous works. To avoid the over-fitting problem brought by the method proposed in~\cite{cheng-etal-2020-spellgcn} that continues to fine-tune the trained model on the SIGHAN13 training data before testing, we follow the post-processing method implemented in~\cite{xu-etal-2021-read, li-etal-2022-past} and don't consider all the detected/corrected mixed auxiliary, which will not compromise the fairness of our experiments and can better reflect the model's real performance.

\end{document}